# NAVIGATIONAL RULE DERIVATION: AN ALGORITHM TO DETERMINE THE EFFECT OF TRAFFIC SIGNS ON ROAD NETWORKS[*]

Daniil Galaktionov, Laboratorio de Bases de Datos, Facultad de Informática, Universidade da Coruña, 15071 A Coruña, Spain, d.galaktionov@udc.es

Miguel R. Luaces, Laboratorio de Bases de Datos, Facultad de Informática, Universidade da Coruña, 15071 A Coruña, Spain, luaces@udc.es

Ángeles S. Places, Laboratorio de Bases de Datos, Facultad de Informática, Universidade da Coruña, 15071 A Coruña, Spain, asplaces@udc.es

## Abstract

*In this paper we present an algorithm to build a road network map enriched with traffic rules such as one-way streets and forbidden turns, based on the interpretation of already detected and classified traffic signs. Such algorithm helps to automatize the elaboration of maps for commercial navigation systems.*

*Our solution is based on simulating navigation along the road network, determining at each point of interest the visibility of the signs and their effect on the roads. We test our approach in a small urban network and discuss various ways to generalize it to support more complex environments.*

---

[*] Funded in part by European Union's Horizon 2020 research and innovation programme under the Marie Sk lodowska-Curie grant agreement No 690941, Ministerio de Economía y Competitividad (PGE y Fondos FEDER) ref. TIN2013-46801-C4-3-R; CDTI and Ministerio de Economía y Competitividad, Ref. IDI-20141259, Ref. ITC-20151305, and Ref. ITC-20151247; and by Xunta de Galicia (co-founded with FEDER) [GRC2013/053].

# 1 INTRODUCTION

Any commercial GPS navigation system provides a routing feature that informs the user on routes, formed by possible paths over a road network, to reach a destination point from a given origin. The calculations involved are expected to produce routes with certain soft warranties about them, like being the quickest possible routes or sometimes avoiding traffic congestions. There are also more important strict warranties, most notably that a route is physically and legally navigable. For these strict warranties, the system must be supplied with up-to-date representations of the road network and all the rules associated to it, like one-way streets and turn restrictions. Some of these rules, like speed limits, are also useful for providing soft warranties.

To this day the elaboration of these enriched network maps has been done using semi-automatic techniques. Modern automatic systems exist for the mapping of navigable roads (Cao & Krumm 2009) and also the inventory of traffic signs (Brkić et al. 2009; Maldonado-Bascon et al. 2008). However, this automatic acquisition must always be followed by a manual process of enrichment with the identification of lanes over roads, mapping of traffic rules, road names and other features of interest to make a useful navigational map. In this article we offer a practical algorithm called the *Navigational Rule Derivation (NRD)* that assists in the mapping of traffic rules on a road network, alleviating the amount of manual labour needed, thus reducing the cost of producing and updating these maps.

To this date, no existing solution has been found for this specific problem. However, there is a big focus on research for sign detection and classification systems, most notably (Mogelmose et al. 2012) and (Cireşan et al. 2012), among many others. The techniques developed by these authors are very helpful for traffic inventory, as they allow to automatically collect large quantities of data about signs, their visibility and their positions. The current work plays the role of a possible next step for these systems, which would be establishing the effect that these signs have on the traffic network. To do this, the signs must be classified and their positions and orientations need to be precisely extracted. This may be developed using LIDAR systems as done in (Tagaki et al. 2006), (Zhou & Deng 2014) and more recently in (Riveiro et al. 2016).

The remainder of this article is organized as follows. In Section 2 we give a complete overview of the data we work with and our goals. Our work is described in two sections. In Section 3, we offer the final algorithm and analyze its results in the test playground, while in Section 4 we describe some interesting alternative solutions that were discarded due to their unsuccessful performance. Finally, Section 5 gives the conclusions.

# 2 PROBLEM DESCRIPTION

There are two kinds of prepared information that we part from: a set of detected and classified signs that is combined with a navigable road network. We store some useful information about the signs: their positions, their types and their orientations relative to the north, also called azimuths. This is all the data we need to define our problem, being our main goal to automatically produce an enriched version of this network with traffic rules extracted from this information, so it can be used to make better routing algorithms. As a desirable side effect, our rules may be published conforming to a standard, such as the INSPIRE Data Specification on Transport Network[†].

---

[†] http://inspire.jrc.ec.europa.eu/documents/Data_Specifications/INSPIRE_DataSpecification_TN_v3.2.pdf

| | | | |
|---|---|---|---|
| 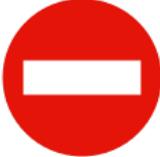 | **R-101**<br>**No way**<br>Forbids accessing a road from where this sign can be seen. | 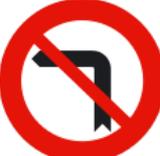 | **R-400a**<br>**One way to the right**<br>Forces vehicles to follow a direction. |
| 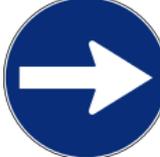 | **R-302**<br>**No right turn**<br>Forbids turning to the right. | 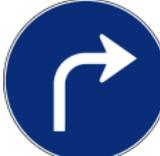 | **R-400b**<br>**One way to the left**<br>Forces vehicles to follow a direction. |
| 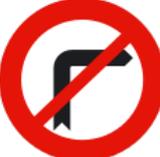 | **R-303**<br>**No left turn**<br>Forbids turning to the left. | 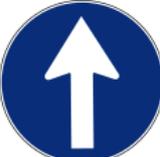 | **R-400d**<br>**Turn right**<br>Mandatory turn to the right. |
| 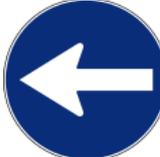 | **R-400c**<br>**Drive ahead**<br>Restricts any possible turns at the sign's position. | 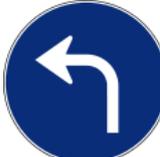 | **R-400e**<br>**Turn left**<br>Mandatory turn to the left. |

*Table 1.     Mapped signs and their descriptions.*

We have decided to simplify the problem for our approach, so we have limited our algorithm to work with just a set of sign types, described in Table 1. Given that the algorithm uses the type of sign instead of its image, it can be applied to any country where similar signs exist. Furthermore, even though we present the Spanish version of the sign, these signs are used all over Europe and many other countries. We also force ourselves to work with a model of the traffic world that makes the following assumptions:

- For every road there are at most two lanes, one for each navigable direction.
- The detected traffic signs alone are enough to infer all the legal restrictions for a road network. This also means that any other sources of information like road markings can be safely ignored.
- There is no other information needed to correctly assign rules to roads, so everything is located at the same height or level and there are neither buildings nor other obstacles for the visibility of a sign.
- The signs **R-101**, **R-400a**, **R-400b** and **R-400c** are found at intersections or just before entering a street and the signs **R-302**, **R-303**, **R-400d** and **R-400e** are found before reaching an intersection, at a reasonable distance of about ten meters.
- Some signs are complimentary.
    - For instance, an R-400a sign would have the same effect if instead there was an R-101 sign for every street from the intersection except the one to the right of the sign.
    - There is a similar effect for R-400d and R-400e, which could be replaced by turn restrictions to every other street.

# 3 THE NAVIGATIONAL RULE DERIVATION ALGORITHM

## 3.1 Overview

The idea behind the NRD is to mimic the natural behaviour of human drivers while they drive through a relatively unknown road network, such as a city. The drivers must take notice of the signs while navigating the network, this way discovering where they may navigate from their current vehicle position and effectively mapping rules to streets within vision.

We may safely assume that the rules from all signs that restrict navigation can be mapped when an intersection is reached after navigating a street. While no intersection is reached, there is no decision to be made, as sudden U-turns are usually prohibited. But as mentioned in Section 2, there are signs that involve navigation restrictions and placed before the intersection, at any point along the street, so it is important to detect and *remember* these kind of signs until an intersection is reached, just as a human driver would do.

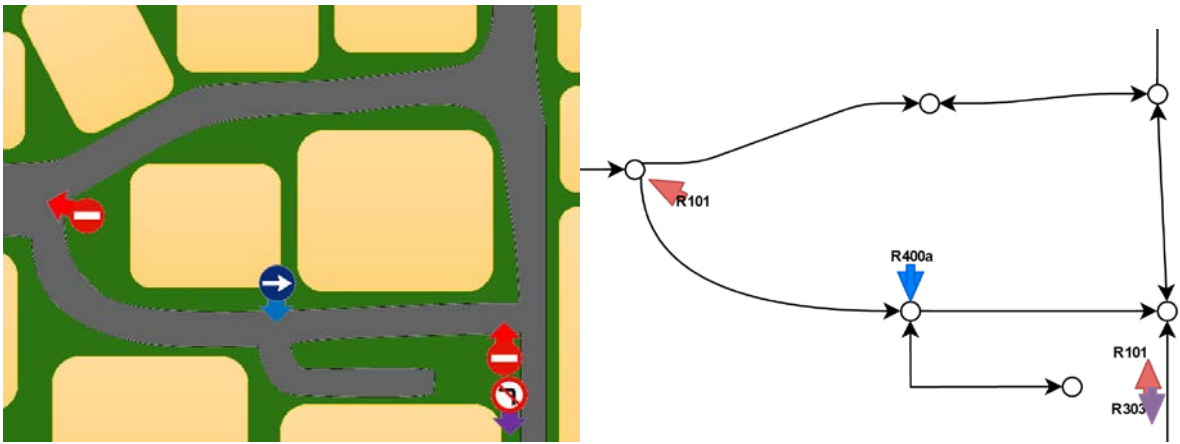

*Figure 1.*   (left) A small sample of a typical urban street layout with signs and their azimuths. (right) Its equivalent representation in terms of directed edges and nodes, preserving the signs.

From this point, we will use the terms **edge** and **node** from graph terminology. A node is a traffic intersection, that is, any point where a turn decision is possible. A node has usually three or more outgoing edges, but it can also be a dead end, with no other outgoing edges. An edge is any navigable link between two nodes. A typical city street is split into multiple edges that connect its intersections with other streets (nodes), as shown in Figure 1. All edges are directed, meaning that they can be navigated in only one way, from node A to node B. A different edge is needed to reciprocally connect B to A.

The algorithm depicted in Figure 2 starts with a graph representation for the traffic network of interest, with every road split into directed edges for both directions, since we initially must consider that every road is always navigable from any direction. The driving starts from a single edge, detecting all signs along this edge that are oriented for the current direction, and also all signs that can be seen from the node in the end. When a node is reached, each of the previously detected signs is analyzed to create corresponding rules, as explained in Section 3.3. All the outgoing edges from this node that do not violate any of these rules are then added into a queue called **the frontier**, and marked as visited so they would never be added more than once. The first item of the frontier is then extracted and navigated as before. The algorithm ends when the frontier is empty, which means it cannot navigate any further. If the graph had isolated groups of edges, they would be never visited[‡].

---

*assignSigns* *(graph, startEdge):*
    *frontier* ← **empty Queue**;
    *frontier.***push***(startEdge);*

    **while** *frontier is not empty:*
        *currentEdge* ← *frontier.***pop***();*
        *node* ← *currentEdge.destination;*
        *outgoingEdges* ← *node.outgoingEdges;*

        *signs* ← **detectSignsAlong***(currentEdge);*
        *signs* ← *signs* ∪ **detectSignsFrom***(node);*
        *rules* ← **analyzeSigns***(signs, currentEdge, node, outgoingEdges, frontier);*

        **for each** *edge* **in** *outgoingEdges:*
            **if** *not edge.visited*
                **and** *not* **isNavigationForbidden** *(currentEdge, edge, rules):*
                    *edge.visited* ← *True;*
                    *frontier.***push***(edge);*

*Figure 2.*     *A high-level view of the NRD.*

Two checks are performed in **isNavigationForbidden** from Figure 2 that could avoid traversing an edge:

- There is a rule that forbids driving through *edge*, as would usually happen for one-way streets or when turn restrictions exist from *currentEdge* to *edge*.

- *edge* is the opposite for *currentEdge*, as sudden U-turns are not usually legal. An exception is made when it is the only available edge left, whether it is because *node* is a dead end or because all the other edges from *outgoingEdges* were discarded by some of the previous checks. Making this exception allows the NRD to fully cover all the navigable roads in both directions.

In the following section we will proceed to explain the working of **detectSignsFrom** and **detectSignsAlong**.

### 3.2     Sign detection

The goal of this stage is to determine what signs can be seen while navigating the current edge and upon reaching the intersection node. As previously implied, two different detectors are needed to effectively find all the visible signs. The detector described in Figure 3 returns visible signs from an intersection node. First it obtains all signs that are at a distance of 15m and then it discards the ones that are not oriented towards the intersection. To achieve a reasonable temporal efficiency, we implement the **filter** operation using a spatial index such as an R-Tree (Guttman 1984). Another operation involved in this detector is **calculateHeading**, which returns the bearing of point B from point A relative to the north[§]. In case the sign appears to be rotated by more than 80º from the position of the driver, we consider that it is not meant to be visible from the intersection.

---

[‡] In theory, it does not make sense to derive traffic rules for roads that can never be reached. In practice, it could be done by making artificial edges with no geometry that connect isolated edge groups.

---

*detectSignsFrom* (node):
    *signsOfInterest* ← *{R-101, R-400a, R-400b, R-400c}*;
    *detectedSigns* ← **filter** *every sign within 15m from node*;

    ***for each*** *sign* ***in*** *detectedSigns*:
        *signHeading* ← **calculateHeading**(*myPosition, sign*);

        ***if*** *sign.type* ∉ *signsOfInterest*:
            ***or*** **abs**(**normalize**(*signHeading* + *sign.azimut*)) > 80º:
                *detectedSigns*.**remove**(*sign*);

---

*Figure 3.*     *From node sign detector.*

A slightly more complex variant is used for the second detector in Figure 4, where we make use of length-indexed lines. An operation called **project** allows returning the coordinates of a length within a line (e.g., its 24th meter). Its complimentary operation is **index**, which returns the length of a line up to a given point. Finally, the **closest** operation returns the coordinates of the closest point from a line to another given point. Instead of using always a fixed position to figure out the visibility of a sign, we take the closest position of the edge to the sign, and then use a point that is ten meters behind it along the edge as the reference position. Notice that when the closest point is either the start or the end of the edge, then it is discarded as it should be detected from an intersection, if ever.

---

*detectSignsAlong* (currentEdge):
    *signsOfInterest* ← *{R-302, R-303, R-400d, R-400e}*;
    *detectedSigns* ← **filter** *every sign within 10m from currentEdge*;

    ***for each*** *sign* ***in*** *detectedSigns*:
        ***if*** *sign.type* ∉ *signsOfInterest*:
            *detectedSigns*.**remove**(*sign*);
        ***else***:
            *signProj* ← **index**(**closest**(*currentEdge, sign*));
            *myPosition* ← **project**(*currentEdge*, **max**(*0, signProj - 10m*));
            *signHeading* ← **calculateHeading**(*myPosition, sign*);

            ***if*** *signProj* ≤ *0*
                ***or*** *signProj* ≥ **length**(*currentEdge*):
                ***or*** **abs**(**normalize**(*signHeading* + *sign.azimut*)) > 80º:
                    *detectedSigns*.**remove**(*sign*);

---

*Figure 4.*     *Along edge sign detector.*

---

[§] 0º when B is to the north of A, 90º when B is to the east, etc...

## 3.3 Rule generation

For each detected sign, a score is calculated for every outgoing edge with a likelihood of this sign applying to it. If the edge with the best score is above a threshold of zero, then a new rule is generated. This, however, may lead to a problem when a sign can be detected from more than one edge or node due to their proximity and generating spurious rules, as depicted in Figure 5. To solve it, the newly generated rule is associated to the sign they were generated from, and can only be replaced when a higher score is obtained. The unvisited edges affected by the replaced rules may be never visited naturally, so they have to be pushed into the frontier, as shown in Figure 7.

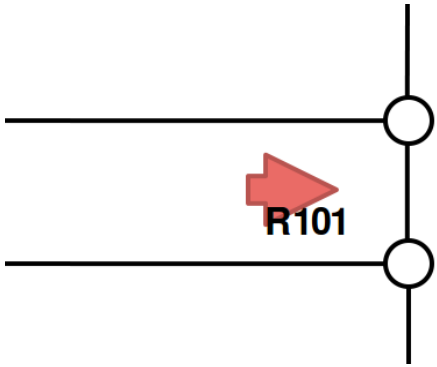

*Figure 5.*   A simple example where a sign would be detected from more than one node. In this case, the lower horizontal edge would have a higher score, as it is to the left of the sign.

---

**analyzeSigns** *(signs, currentEdge, node, outgoingEdges, frontier):*
    *noWaySigns* ← *{R-101};*
    *noTurnSigns* ← *{R-302, R-303};*
    *oneWaySigns* ← *{R-400a, R-400b, R-400c};*
    *mustTurnSigns* ← *{R-400d, R-400e};*
    *newRules* ← *{};*

    **for each** *sign* **in** *signs* :
        **if** *sign.type* ∈ *noWaySigns:*
            *bestEdge, score* ← **bestNoWayEdge***(sign, node, outgoingEdges);*
            **associateNewRule***(sign,* **NoWayRule***(bestEdge),*
                *score, True, frotiner);*
        **else if** *sign.type* ∈ *noT urnSigns:*
            *bestEdge, score* ← **bestNoTurnEdge***(sign, currentEdge, outgoingEdges);*
            **associateNewRule***(sign,* **NoTurnRule***(currentEdge, bestEdge),*
                *score, False, frontier);*
        **else if** *sign.type* ∈ *oneWaySigns:*
            *(omitted)*
        **else if** *sign.type* ∈ *mustTurnSigns:*
            *(omitted)*

    *newRules* ← *newRules* ∪ *{sign.rule};*

    **return** *newRules;*

---

*Figure 6.*   Generating rules from detected signs.

In Figure 6, the signs that have one of the **oneWaySigns** types can replaced by various **NoWayRules**, according to what is mentioned in Section 2 or a special rule that contains a set of banned edges instead of a single banned edge. In a similar way, **mustTurnSigns** produce rules that are effectively **NoTurnRules** with one source edge and a set of destination edges. They are omitted in pseudocode as their cases would need a more complex and generic **associateNewRule** routine that the one shown in Figure 7.

---

*associateNewRule* (sign, newRule, score, unban, frontier):
    edge ← sign.rule.edge;
    **if** score > 0:
        **if** ∃ sign.rule **and** score > sign.rule.score:
            **if** unban:
                edge.banned ← False;

            **if** not edge.visited **and** not edge.banned:
                edge.visited ← True;
                frontier.**push**(edge);

        sign.score ← score;
        sign.rule ← newRule;

---

Figure 7.     associateNewRule routine.

Pseudocode for the calculation of the edge scores can be found in Figures 8 and 9. It uses length-indexed lines explained in Section 3.2 and also the **angle** operation shown in Figure 10, that calculates the smallest angle between two vectors defined as three points: the first and the third parameters are the tips of each vector, while the second is a common tail for both of them, returning a value between -180º and 180º.

---

*bestNoWayEdge* (sign, node, outgoingEdges):
    score ← -∞;

    **for each** edge **in** outgoingEdges:
        signProj ← **index**(**closest**(edge, sign));
        **if** signProj ≤ 0:
            signProj ← 10m;

        α ← **angle**(sign, node, **project**(edge, signProj));
        **if** α < -10º:
            α ← α - 30º;

        curScore ← 90 - **abs**(α);
        **if** curScore > score:
            score ← curScore;
            bestEdge ← edge;

    **return** bestEdge, score;

---

Figure 8.     Finding the best edge for a R-101 (no way) sign.

```
bestNoTurnEdge (sign, currentEdge, outgoingEdges):
    score ← -∞;
    signProj ← index(closest(currentEdge, sign));
    if signProj ≥ length(currentEdge):
        signProj ← length(currentEdge) - 10m;

    for each edge in outgoingEdges:
        α ← angle(project(currentEdge, signProj), node, index(edge, 10m));
        if sign.type = R-302:
            α ← α – 90º;
        else if sign.type = R-303:
            α ← α + 90º;

        curScore ← 60 - abs(α);
        if curScore > score:
            score ← curScore;
            bestEdge ← edge;

    return bestEdge, score;
```

*Figure 9.    Finding the best edge for a R-302/R-303 (no turn) sign.*

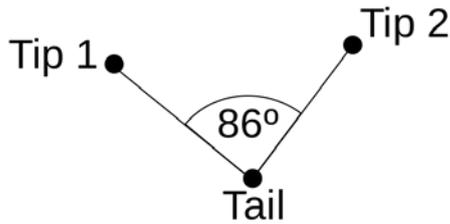

*Figure 10.    Example of the calculation of the relative angle between three points using **angle**(Tip 1, tail, Tip 2) operation.*

In case of **bestNoWayEdge**, we have a node-sign vector, and we search for the most similar node-edge vector, from all the outgoing edges. The least is the angle between both vectors, the higher is the score. We also penalize edges that fall to the right of a sign, as signs are usually put on the same side of a road the traffic drives on, although exceptions are made in some cases. The tip of the node-edge vector is taken as the closest point of the edge to the sign or its 10*th* meter in case the sign is in the opposite direction, which usually means that this is not the edge that the sign applies to.

Similar calculations are made in **bestNoTurnEdge**. This time the vectors are node-currentEdge and node-edge, and we add an offset of -90º of 90º to the angle depending on the direction of the forbidden turn. The closest point of *currentEdge* is used as the tip of the node-currentEdge vector, while the tip of the node-edge vector is always the 10*th* meter of the destination edge.

### 3.4    Playground experimental results

The test playground used to develop our algorithm is located in the Spanish town of Ávila, which is famous for its medieval walls enclosing the old district of the town. This enclosed old district consists of simple one or two-way lane streets, all of them in the same level, and no roundabouts, so it was considered an adequate playground.

The inventory of signs, including their location, orientation and type was obtained following the methodology described in (Riveiro et al. 2016), which works over a point cloud obtained from a LIDAR system. Their approach is based on the sequential application of clustering algorithms to detect traffic signs in the point cloud. These signs are then classified by reconstructing their shape and analyzing their geometric properties.

For the development and tests for our algorithm and all alternative approaches we used PostGIS[**] and JTS[††] library for spatial data storing and manipulation. To evaluate the performance of the NRD in our playground, we crossed the produced data (traffic rules) with already existing data publicly available from OpenStreetMap[‡‡]. Unfortunately, OSM only provides accurate definition of one-way streets but not turn restrictions for Ávila. In fact, most of the turn restrictions are incorrectly mapped as one-way streets, further complicating our validation.

The NRD may fail to extract a correct traffic rule in one of these cases:

1. Real-world signs that were not detected and recognized in the previous stages to our work.
2. Disagreement between the road network and the signs. This is usually caused by having both datasets mapped at different instances of time. Fortunately, the streets from the old part of Ávila are rarely subject to changes.
3. Flaws in the NRD, which probably could be solved using a more complex approach.
4. Bad or incomplete sign placement, as happens in many rural and even urban areas.
5. Oversimplified model. This is not an issue in our playground case, but it is common to find bridges and tunnels in the real world, as well as multiple lanes.

| Rule type | Total mapped rules | Incorrectly placed | Accuracy |
|---|---|---|---|
| One-way streets | 35 | 4 | 88.57% |
| Turn restrictions | 32 | 1 | 96.88% |

*Table 2.      Number of rules generated and their accuracy measures when compared to OSM data.*

In Table 2 we present a measure to the performance of the NRD in terms of accuracy, taking the OSM data and manual observation as ground truth. Notice that we only evaluate direct flaws from the NRD, not its sources of information. Also, it would be difficult to estimate the completeness or "recall" of the NRD, as it is highly dependent on the completeness of the input data itself.

# 4      DISCARDED APPROACHES

## 4.1      By-Sign static mapping

The first temptation was to iterate every sign available and try to figure out which roads it affected to. In theory, this is a much simpler solution as it does not involve any navigation and thus does not require the network to be navigable. A bidimensional cone was projected from the sign, and any affected road had to intersect that cone. An example of two such cones is illustrated in Figure 11

---

[**] http://postgis.net
[††] http://www.vividsolutions.com/jts/JTSHome.htm
[‡‡] http://www.openstreetmap.org

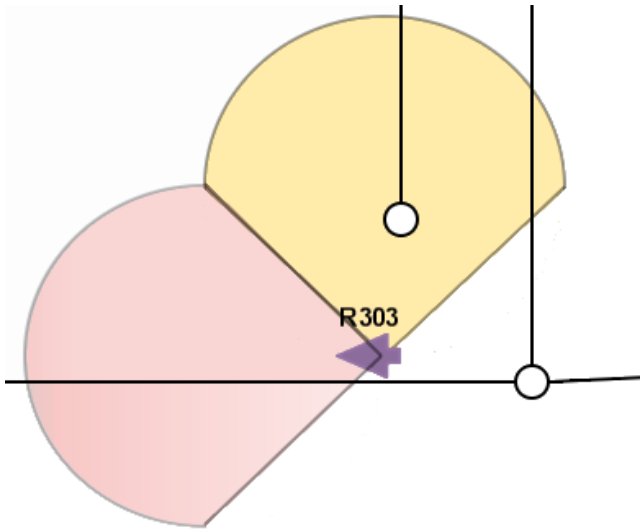

*Figure 11.    Vision cones for a left turn restriction. In the red cone we search for the origin edge, while in the yellow cone we search for the destination.*

. The cone orientation was relative to the sign's azimuth and depended on the sign type. For No Way signs, the cone was opposing the sign's azimuth. For turn restrictions, cones were projected in two different directions to find the affected roads. To choose among various possible matches different scoring methods were experimented with, involving the similarity of the azimuth to the closest segment of the road, distance and relative position. This approach was ultimately proven ineffective for the following reasons:

- It did not take into account the network's connectivity, so the wrong road could be associated. In Figure 11, the yellow cone would detect two edges, disregarding their connectivity, and assign a better score for the bottom edge because it would seem like a closer turn.

- There was no simple way of knowing where a sign was supposed to be seen from, leading to wrong mapping due to this lack of information.

- A sign's azimuth, while vital to calculate its visibility in the NRD, turned out to be a poor source of information to actually derive rules from in the real world.

- This solution cannot be extended to support the mapping of speed limits or roundabouts without resorting to navigation.

### 4.2    Navigation mapping with by-edge decision

In the first versions of the NRD, after navigating into a node, we individually evaluated every outgoing edge by projecting a vision cone towards it. If signs were detected, they were analyzed and the generated rules were associated with that edge. Many spurious rules were generated for signs that could fall into different vision cones, but they could be suppressed by associating them to the sign and using a scoring mechanism as seen in Figure 7. The main issue was that signs of type R-400a, R-400b and R-400c, while meant to be detected from an intersection, were not necessarily close to any outgoing edge it affected, if any at all. A wider cone could have been used, but it made much more sense to detect first every sign within a radius from the intersection first, and then figure out which edge each sign applies to, as finally done for the NRD in Section 3.3.

# 5   CONCLUSIONS

We presented a starting step for a complex problem and provided a valid proof of concept for traffic rule generation algorithms based on the simulation of a virtual driver. From this point, there are numerous enhancements that we could do to the NRD to improve the sign analysis in some complicated cases and to also support more kinds of signs, such as roundabouts and speed limits. When human drivers see a speed limit sign, they remember that limit until they reach some point where the limit is voided, as could be a yield sign or another speed limit. Knowing that, it makes sense to introduce some state into the NRD during navigation, so each edge would be marked with a speed limit as it is visited.

Real-World applications also require a greater degree of complexity than the one we restricted ourselves to in Section 2. We need to support multiple lanes and road markings that could be used to complement or verify rules generated by the usual traffic signs. Also, the navigation needs to work with three-dimensional coordinates or have some information about the level of the roads and signs, which would help to avoid the expected error cases with tunnels and bridges. The NRD can be further enhanced with the addition of trajectory analysis: if we could get data about a significant amount of drivers, we could derive traffic rules by examining turns that nobody takes, their usual driving speed across the network and other useful information.